\documentclass[11pt,a4paper]{article}

\usepackage[hyperref]{paper}
\usepackage{times}
\usepackage{latexsym}
\usepackage{amsmath}
\usepackage{graphicx}
\usepackage{url}
\usepackage{tabularx}
\usepackage{booktabs}
\usepackage{color,soul}

\aclfinalcopy 

\title{Neural Domain Adaptation for Biomedical Question Answering}

\author{Georg Wiese\textsuperscript{1,2}, Dirk Weissenborn\textsuperscript{2} and Mariana Neves\textsuperscript{1} \\
\textsuperscript{1} Hasso Plattner Institute, August Bebel Strasse 88, Potsdam 14482 Germany \\
\textsuperscript{2} Language Technology Lab, DFKI, Alt-Moabit 91c, Berlin, Germany \\
  {\tt georg.wiese@student.hpi.de,}  \\ {\tt  dirk.weissenborn@dfki.de, mariana.neves@hpi.de }
}

\date{}

\begin{document}

\maketitle

\begin{abstract}
  Factoid question answering (QA) has recently benefited from the development of deep learning (DL) systems.
  Neural network models outperform traditional approaches in domains where large datasets exist, such as SQuAD ($\approx100,000$ questions) for Wikipedia articles.
  However, these systems have not yet been applied to QA in more specific domains, such as biomedicine, because datasets are generally too small to train a DL system from scratch.
  For example, the BioASQ dataset for biomedical QA comprises less then $900$ factoid (single answer) and list (multiple answers) QA instances.
  In this work, we adapt a neural QA system trained on a large open-domain dataset (SQuAD, \textit{source}) to a biomedical dataset (BioASQ, \textit{target}) by employing various transfer learning techniques.
  Our network architecture is based on a state-of-the-art QA system, extended with biomedical word embeddings and a novel mechanism to answer list questions.
  In contrast to existing biomedical QA systems, our system does not rely on domain-specific ontologies, parsers or entity taggers, which are expensive to create.
  Despite this fact, our systems achieve state-of-the-art results on factoid questions and competitive results on list questions.
\end{abstract}

\section{Introduction}

Question answering (QA) is the task of retrieving \emph{answers} to a \emph{question} given one or more \emph{contexts}.
It has been explored both in the open-domain setting \cite{voorhees1999trec} as well as domain-specific settings, such as BioASQ for the biomedical domain \cite{tsatsaronis2015overview}.
The BioASQ challenge provides $\approx900$ \emph{factoid} and \emph{list} questions, i.e., questions with one and several answers, respectively.
This work focuses on answering these questions, for example: \textit{Which drugs are included in the FEC-75 regimen? $\rightarrow$ fluorouracil, epirubicin, and cyclophosphamide}.

We further restrict our focus to \emph{extractive} QA, i.e., QA instances where the correct answers can be represented as spans in the contexts.
Contexts are relevant documents which are provided by an information retrieval (IR) system.

Traditionally, a QA pipeline consists of named-entity recognition, question classification, and answer processing steps \cite{salp_qa_chapter}.
These methods have been applied to biomedical datasets, with moderate success \cite{oaqa}.
The creation of large-scale, open-domain datasets such as SQuAD \cite{rajpurkar2016squad} have recently enabled the development of neural QA systems, e.g., \newcite{wang2016machine}, \newcite{dcn}, \newcite{seo2016bidirectional}, \newcite{weissenborn2017fastqa}, leading to impressive performance gains over more traditional systems.

However, creating large-scale QA datasets for more specific domains, such as the biomedical, would be very expensive because of the need for domain experts, and therefore not desirable. The recent success of deep learning based methods on open-domain QA datasets raises the question whether the capabilities of trained models are transferable to another domain via \textit{domain adaptation} techniques. Although domain adaptation has been studied for traditional QA systems \cite{blitzer2007biographies} and deep learning systems \cite{chen2012marginalized,ganin2016domain,bousmalis2016domain,riemer2017forgettingcost,kirkpatrick2017overcoming}, it has to our knowledge not yet been applied for end-to-end neural QA systems.

To bridge this gap we employ various domain adaptation techniques to transfer knowledge from a trained, state-of-the-art neural QA system (FastQA, \newcite{weissenborn2017fastqa}) to the biomedical domain using the much smaller BioASQ dataset.
In order to answer \emph{list questions} in addition to factoid questions, we extend FastQA with a novel answering mechanism.
We evaluate various transfer learning techniques comprehensively.
For factoid questions, we show that mere fine-tuning reaches state-of-the-art results, which can further be improved by a forgetting cost regularization \cite{riemer2017forgettingcost}.
On list questions, the results are competitive to existing systems.
Our manual analysis of a subset of the factoid questions suggests that the results are even better than the automatic evaluation states, revealing that many of the "incorrect" answers are in fact synonyms to the gold-standard answer.

\section{Related Work}

\paragraph{Traditional Question Answering}
Traditional factoid and list question answering pipelines can be subdivided into named-entity recognition, question classification, and answer processing components \cite{salp_qa_chapter}.
Such systems have also been applied to biomedical QA such as the OAQA system by \newcite{oaqa}.
Besides a number of domain-independent features, they incorporate a rich amount of biomedical resources, including a domain-specific parser, entity tagger and thesaurus to retrieve concepts and synonyms.
A logistic regression classifier is used both for question classification and candidate answer scoring.
For candidate answer generation, OAQA employs different strategies for general factoid/list questions, choice questions and quantity questions.

\paragraph{Neural Question Answering}
Neural QA systems differ from traditional approaches in that the algorithm is not subdivided into discrete steps.
Instead, a single model is trained end-to-end to compute an answer directly for a given question and context. The typical architecture of such systems \cite{wang2016machine,dcn,seo2016bidirectional} can be summarized as follows:

\begin{enumerate}
  \item \textbf{Embedding Layer:} Question and context tokens are mapped to a high-dimensional vector space, for example via GloVe embeddings \cite{pennington2014glove} and (optionally) character embeddings \cite{seo2016bidirectional}.
  \item \textbf{Encoding Layer:} The token vectors are processed independently for question and context, usually by a recurrent neural network (RNN).
  \item \textbf{Interaction Layer:} This layer allows for interaction between question and context representations. Examples are Match-LSTM \cite{wang2016machine} and Coattention \cite{dcn}.
  \item \textbf{Answer Layer:} This layer assigns start and end scores to all of the context tokens, which can be done either statically \cite{wang2016machine,seo2016bidirectional} or by a dynamic decoding process \cite{dcn}.
\end{enumerate}

\paragraph{FastQA}
\emph{FastQA} fits into this schema, but reduces the complexity of the architecture by removing the interaction layer, while maintaining state-of-the-art performance \cite{weissenborn2017fastqa}.
Instead of one or several interaction layers of RNNs, FastQA computes two simple word-in-question features for each token, which are appended to the embedding vectors before the encoding layer.
We chose to base our work on this architecture because of its state-of-the-art performance, faster training time and reduced number of parameters.

\paragraph{Unsupervised Domain Adaptation}
Unsupervised domain adaptation describes the task of learning a predictor in a \emph{target domain} while labeled training data only exists in a different \emph{source domain}.
In the context of deep learning, a common method is to first train an autoencoder on a large unlabeled corpus from both domains and then use the learned input representations as input features to a network trained on the actual task using the labeled source domain dataset \cite{glorot2011domain,chen2012marginalized}.
Another approach is to learn the hidden representations directly on the target task.
For example, domain-adversarial training optimizes the network such that it computes hidden representations that both help predictions on the source domain dataset and are indistinguishable from hidden representations of the unlabeled target domain dataset \cite{ganin2016domain}.
These techniques cannot be straightforwardly applied to the question answering task, because they require a large corpus of biomedical question-context pairs (albeit no answers are required).

\paragraph{Supervised Domain Adaptation}

In contrast to the unsupervised case, supervised domain adaptation assumes access to a small amount of labeled training data in the target domain.
The simplest approach to supervised domain adaptation for neural models is to pre-train the network on data from the source domain and then fine-tune its parameters on data from the target domain.
The main drawback of this approach is \emph{catastrophic forgetting}, which describes the phenomenon that neural networks tend to "forget" knowledge, i.e., its performance in the source domain drops significantly when they are trained on the new dataset. Even though we do not directly aim for good performance in the source domain, measures against catastrophic forgetting can serve as a useful regularizer to prevent over-fitting.

Progressive neural networks combat this issue by keeping the original parameters fixed and adding new units that can access previously learned features \cite{rusu2016progressive}.
Because this method adds a significant amount of new parameters which have to be trained from scratch, it is not well-suited if the target domain dataset is small.
\newcite{riemer2017forgettingcost} use fine-tuning, but add an additional \emph{forgetting cost} term that punishes deviations from predictions with the original parameters.
Another approach is to add an L2 loss which punishes deviation from the original parameters.
\newcite{kirkpatrick2017overcoming} apply this loss selectively on parameters which are important in the source domain.

\section{Model}

\begin{figure}[h]
\label{fig:biomedical_qa}
\centering
\includegraphics[width=0.45\textwidth]{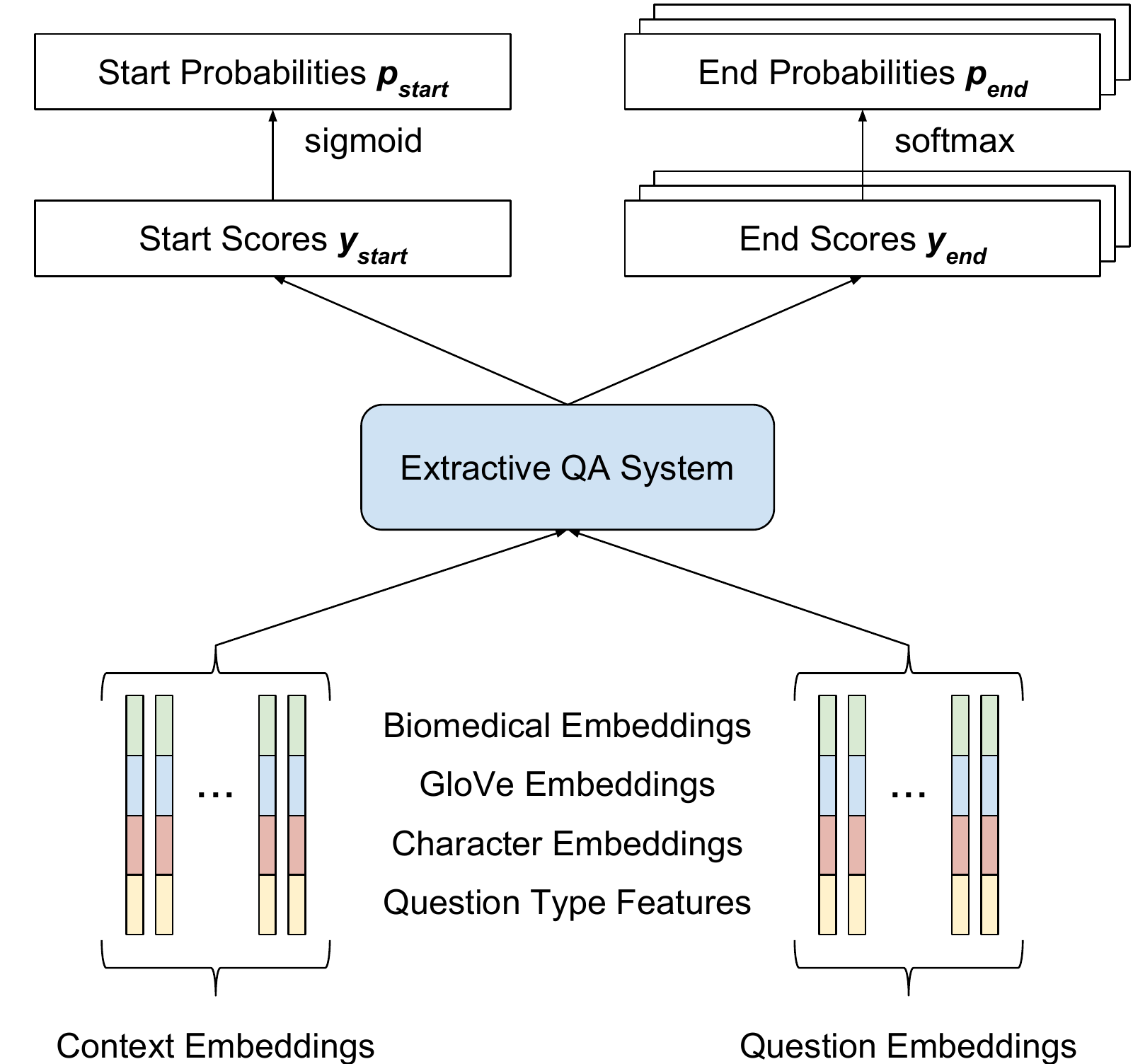}
\caption{
Network architecture of our system for biomedical question answering.
At its core, it uses an extractive neural QA system as a black box (we use FastQA \protect\cite{weissenborn2017fastqa}).
The embedding layer is modified in order to include biomedical word embeddings and question type features.
The output layer is adjusted to add the ability to answer list questions in addition to factoid questions.
}
\end{figure}

Our network architecture is based on FastQA \cite{weissenborn2017fastqa}, a state-of-the-art neural QA system.
Because the network architecture itself is exchangeable, we treat it as a black box, with subtle changes at the input and output layer as well as to the decoding and training procedure. These changes are described in the following.
See Figure~\ref{fig:biomedical_qa} for an overview of the system.

\subsection{Input Layer}
\label{sec:input_layer}

In a first step, words are embedded into a high-dimensional vector space.
We use three sources of embeddings, which are concatenated to form a single embedding vector:

\begin{itemize}

\item
\textbf{GloVe embeddings:} 300-dimensional GloVe vectors \cite{pennington2014glove}. These are open-domain word vectors trained on 840 billion tokens from web documents. The vectors are not updated during training.

\item
\textbf{Character embeddings:} As used in FastQA \cite{weissenborn2017fastqa} and proposed originally by \newcite{seo2016bidirectional}, we employ a 1-dimensional convolutional neural network which computes word embeddings from the characters of the word. 
\item
\textbf{Biomedical Word2Vec embeddings:} 200-dimensional vectors trained using Word2Vec \cite{mikolov2013distributed_word2vec} on about 10 million PubMed abstracts \cite{biomedical_word2vec}. These vectors are specific to the biomedical domain and we expect them to help on biomedical QA.

\end{itemize}

As an optional step, we add entity tag features to the token embeddings via concatenation.
Entity tags are provided by a dictionary-based entity tagger based on the UMLS Metathesaurus.
The entity tag feature vector is a $127$-dimensional bit vector that for each of the UMLS semantic types states whether the current token is part of an entity of that type.
This step is only applied if explicitly noted.

Finally, a one-hot encoding of the question type (\emph{factoid} or \emph{list}) is appended to all the input vectors.
With these embedding vectors as input, we invoke FastQA to produce start and end scores for each of the $n$ context tokens.
We denote start scores by $y_{start}^{i}$ and end scores conditioned on a predicted start at position $i$ by $y_{end}^{i, j}$, with start index $i \in [1, n]$ and end index $j \in [i, n]$.

\subsection{Output Layer}

In our adapted output layer, we convert the start and end scores to span probabilities.
The computation of these probabilities is independent of the question type.
The interpretation, however, depends on the question type:
While for factoid questions, the list of answer spans is interpreted as a ranked list of answer candidates, for list questions, answers above a certain probability threshold are interpreted as the set of answers to the question.

Given the start scores $y_{start}^1, ..., y_{start}^n$ and end scores $y_{end}^{i, 1}, ..., y_{end}^{i, n}$, we compute the start and end probabilities as follows:

\begin{equation}
\label{eq:p_s}
  p_{start}^i = \sigma(y_{start}^i)
\end{equation}

\begin{equation}
\label{eq:p_e}
  p_{end}^{i, \cdot} = \operatorname{softmax}(y_{end}^{i, \cdot})
\end{equation}

\noindent
where $\sigma(x)$ is the sigmoid function.
As a consequence, multiple tokens can be chosen as likely start tokens, but the network is expected to select a single end token for a given start token, hence the $\operatorname{softmax}$ function.
Finally, the probability that a given span $(i, j)$ answers the question is $p_{span}^{i, j} = p_{start}^{i} \cdot p_{end}^{i, j}$.
This extension generalizes the FastQA output layer such that multiple answer spans with different start positions can have a high probability, allowing us to retrieve multiple answers for list questions.

\subsection{Decoding}
Given a trained model, start probabilities can be obtained by running a forward pass and computing the start probability as in Equation~\ref{eq:p_s}.
For the top $20$ starts, we compute the end probabilities as given by Eq.~\ref{eq:p_e}.
From the start and end probabilities, we extract the top $20$ answer spans ranked by $p_{span}^{i, j}$.
As a simple post-processing step, we remove duplicate strings and retain only those with the highest probability. 

For factoid questions, we output the $5$ most likely answer spans as our ranked list of answers.
For list questions, we learn a \emph{probability cutoff threshold} $t$ that defines the set of list answers $A = \{(i, j) | p_{span}^{i, j} \ge t\}$.
We choose $t$ to be the threshold that optimizes the list F1 score on the respective development set.

\subsection{Domain Adaptation}
\label{sec:optimization_da}

\paragraph{Fine-tuning}
Our training procedure consists of two phases:
In the \emph{pre-training phase}, we train the model on SQuAD, using a token F1 score as the training objective as by \newcite{weissenborn2017fastqa}.
We will refer to the resulting parameters as the \emph{base model}.
In the \emph{fine-tuning phase}, we initialize the model parameters with the base model and then continue our optimization on the BioASQ dataset with a smaller learning rate.

\paragraph{Forgetting Cost Regularization}
To avoid \textit{catastrophic forgetting} during fine-tuning as a means to regularize our model, we optionally add an additional forgetting cost term $L_{fc}$, as proposed by \newcite{riemer2017forgettingcost}.
It is defined as the cross-entropy loss between the current predictions and the base model's predictions.

\paragraph{L2 Weight Regularization} We also add an L2 loss term $L_{l2}$ which penalizes deviations from the base model's parameters.
Note that a more advanced approach would be to apply this loss selectively on weights which are particularly important in the source domain \cite{kirkpatrick2017overcoming}.
The final loss is computed as $L_{final} = L_{original} + C_{fc} \cdot L_{fc} + C_{l2} \cdot L_{l2}$ where $C_{fc}$ and $C_{l2}$ are hyperparameters which are set to $0$ unless otherwise noted.

\section{Experimental Setup}

\subsection{Datasets}

\paragraph{SQuAD}
\emph{SQuAD} \cite{rajpurkar2016squad} is a dataset of $\approx100,000$ questions with relevant contexts and answers that sparked research interest into the development of neural QA systems recently.
The contexts are excerpts of Wikipedia articles for which crowd-source workers generated questions-answer pairs. Because of the large amount of training examples in SQuAD, it lends itself perfectly as our source dataset.

\paragraph{BioASQ}
The BioASQ challenge provides a biomedical QA dataset \cite{tsatsaronis2015overview} consisting of questions, relevant contexts (called \emph{snippets}) from PubMed abstracts and possible answers to the question. It was carefully created with the help of biomedical experts.

In this work, we focus on Task B, Phase B of the BioASQ challenge, in which systems must answer questions from gold-standard snippets.
These questions can be either yes/no questions, summary questions, factoid questions, or list questions.
Because we employ an extractive QA system, we restrict this study to answering factoid and list questions by extracting answer spans from the provided contexts.

The 2017 BioASQ training dataset contains $1,799$ questions, of which $413$ are factoid and $486$ are list questions.
The questions have $\approx20$ snippets on average, each of which are on average $\approx34$ tokens long.
We found that around $65\%$ of the factoid questions and around $92\%$ of the list questions have at least one extractable answer.
For questions with extractable answers, answers spans are computed via a simple substring search in the provided snippets.
All other questions are ignored during training and treated as answered incorrectly during evaluation.

\subsection{Training}
\label{sec:evaluation_training}
We minimize the cross-entropy loss for the gold standard answer spans.
However, for multiple answer spans that refer to the same answer (e.g. synonyms), we only minimize the loss for the span of the lowest loss.
We use the ADAM \cite{kingma2014adam} for optimization on SQuAD with a learning rate starting at $10^{-3}$ which is halved whenever performance drops between checkpoints. During the fine-tuning phase, we continue optimization on the BioASQ dataset with a smaller learning rate starting at $10^{-4}$.
During both phases, the model is regularized by variational dropout of rate $0.5$ \cite{gal2015dropout}.

\subsection{Evaluation}

The official evaluation measures from BioASQ are mean reciprocal rank (MRR) for factoid questions and F1 score for list questions \footnote{The details can be found at \url{http://participants-area.bioasq.org/Tasks/b/eval_meas/}}.
For factoid questions, the list of ranked answers can be at most five entries long.
The F1 score is measured on the gold standard list elements.
For both measures, case-insensitive string matches are used to check the correctness of a given answer.
A list of synonyms is provided for all gold-standard answers.
If the system's response matches one of them, the answer counts as correct.

For evaluation, we use two different fine-tuning datasets, depending on the experiment:
\emph{BioASQ3B}, which contains all questions of the first three BioASQ challenges, and \emph{BioASQ4B} which additionally contains the test questions of the fourth challenge.
\emph{BioASQ4B} is used as the training dataset for the fifth BioASQ challenge whereas \emph{BioASQ3B} was used for training during the fourth challenge.

Because the datasets are small, we perform 5-fold cross-validation and report the average performance across the five folds.
We use the larger \emph{BioASQ4B} dataset except when evaluating the ensemble and when comparing to participating systems of previous BioASQ challenges.

All models were implemented using TensorFlow \cite{abadi2016tensorflow} with a hidden size of $100$.
Because the context in BioASQ usually comprises multiple snippets, they are processed independently in parallel for each question. Answers from all snippets belonging to a question are merged and ranked according to their individual probabilities.

\section{Results}

\subsection{Domain Adaptation}

\begin{table*}[t]
  \centering
\begin{tabularx}{\textwidth}{X l l}
\toprule
  \textbf{Experiment} & \textbf{Factoid MRR} & \textbf{List F1} \\
\midrule
(1) Training on BioASQ only & 17.9\% & 19.1\% \\
\midrule
(2) Training on SQuAD only & 20.0\% & 8.1\% \\
(3) Fine-tuning on BioASQ & 24.6\% & 23.6\% \\
\midrule
(4) Fine-tuning on BioASQ w/o biomedical embeddings & 21.3\% & 22.4\% \\
(5) Fine-tuning on BioASQ w/ entity features & 23.3\% & 23.8\% \\
\midrule
(6) Fine-tuning on BioASQ + SQuAD & 23.9\% & 23.8\% \\
(7) Fine-tuning on BioASQ w/ forgetting cost ($C_{fc} = 100.0$) & 26.2\% & 21.1\% \\
(8) Fine-tuning on BioASQ w/ L2 loss on original parameters ($C_{l2} = 0.3$) & 22.6\% & 20.4\% \\
\bottomrule
\end{tabularx}
  \caption{
  Comparison of various transfer learning techniques.
  In Experiment 1, the model was trained on BioASQ only.
  In Experiment 2, the model was trained on SQuAD and tested on BioASQ.
  We refer to it as the \emph{base model}.
  In Experiment 3, the base model parameters were fine-tuned on the BioASQ training set.
  Experiments 4-5 evaluate the utility of domain dependent word vectors and features.
  Experiments 6-8 address the problem of catastrophic forgetting.
  All experiments have been conducted with the \emph{BioASQ4B} dataset and 5-fold cross-validation.
  }
  \label{tab:results_transfer_learning}
\end{table*}

In this section, we evaluate various domain adaptation techniques.
The results of the experiments are summarized in Table~\ref{tab:results_transfer_learning}.

\paragraph{Baseline}
As a baseline without transfer learning, Experiment 1 trains the model on BioASQ only.
Because the BioASQ dataset by itself is very small, a dropout rate of $0.7$ was used, because it worked best in preliminary experiments.
We observe a rather low performance, which is expected when applying deep learning to such a small dataset.

\paragraph{Fine-tuning}
Experiments 2 and 3 evaluate the pure fine-tuning approach:
Our \emph{base model} is a system trained on SQuAD only and tested on BioASQ (Experiment 2).
For Experiment 3, we fine-tuned the base model on the \emph{BioASQ4B} training set.
We observe that performance increases significantly, especially on list questions.
This increase is expected, because the network is trained on biomedical- and list questions, which are not part of the SQuAD dataset, for the first time.
Overall, the performance of the fine-tuned model on both question types is much higher than the baseline system without transfer learning.

\paragraph{Features}
\label{sec:results_features}
In order to evaluate the impact of using biomedical word embeddings, we repeat Experiment 3 without them (Experiment 4).
We see a factoid and list performance drop of $3.3$ and $1.2$ percentage points, respectively, showing that biomedical word embeddings help increase performance.

In Experiment 5, we append entity features to the word vector, as described in Section~\ref{sec:input_layer}.
Even though these features provide the network with domain-specific knowledge, we found that it actually harms performance on factoid questions.
Because most of the entity features are only active during fine-tuning with the small dataset, we conjecture that the performance decrease is due to over-fitting.

\paragraph{Catastrophic Forgetting}

We continue our study with techniques to combat catastrophic forgetting as a means to regularize training during fine-tuning.
In Experiment 6 of Table~\ref{tab:results_transfer_learning} we fine-tune the base model on a half-half mixture of BioASQ and SQuAD questions (BioASQ questions have been upsampled accordingly).
This form of joint training yielded no significant performance gains.
Experiment 7 regularizes the model via an additional forgetting cost term, as proposed by \newcite{riemer2017forgettingcost} and explained in Section~\ref{sec:optimization_da}.
We generally found that this technique only increases performance for factoid questions where the performance boost was largest for $C_{fc} = 100.0$.
The fact that the forgetting loss decreases performance on list questions is not surprising, as predictions are pushed more towards the predictions of the base model, which has very poor performance on list questions.

Experiment 8 adds an L2 loss which penalizes deviations from the base model's parameters.
We found that performance decreases as we increase the value of $C_{l2}$ which shows that this technique does not help at all. For the sake of completeness we report results for $C_{l2} = 0.3$, the lowest value that yielded a significant drop in performance.

\subsection{Ensemble}
\label{sec:ensemble}

Model ensembles are a common method to tweak the performance of a machine learning system.
Ensembles combine multiple model predictions, for example by averaging, in order to improve generalization and prevent over-fitting.
We evaluate the utility of an ensemble by training five models on the \emph{BioASQ3B} dataset using 5-fold cross-validation.
Each of the models is evaluated on the 4B test data, i.e., data which is not included in \emph{BioASQ3B}.

During application, we run an ensemble by averaging the start and end scores of individual models before they are passed to the sigmoid / softmax functions as defined in Eq.~\ref{eq:p_s} and~\ref{eq:p_e}.
In Table~\ref{tab:results_ensemble} we summarize the average performance of the five models, the best performance across the five models, and the performance of the ensemble.
We observe performance gains of $3$ percentage points on factoid questions and a less than $1$ percentage point on list questions, relative to the best single model.
This demonstrates a small performance gain that is consistent with the literature.

\begin{table}
  \centering
\begin{tabular}{l l l}
\toprule
  \textbf{Experiment} & \textbf{Factoid MRR} & \textbf{List F1} \\
\midrule
Average & 23.4\% & 24.0\% \\
Best & 24.3\% & 27.7\% \\
Ensemble & 27.3\% & 28.6\% \\
\bottomrule
\end{tabular}
  \caption{
  Performance of a model ensemble.
  Five models have been trained on the \emph{BioASQ3B} dataset and tested on the 4B test questions.
  We report the average and best single model performances, as well as the ensemble performance.}
  \label{tab:results_ensemble}
\end{table}

\subsection{Comparison to competing BioASQ systems}

\begin{table*}[t]
  \centering
\begin{tabularx}{\textwidth}{l | l l l | l l l}
\toprule
\multicolumn{1}{c|}{} & \multicolumn{3}{c|}{Factoid MRR} & \multicolumn{3}{c}{List F1} \\
  \textbf{Batch} & \textbf{Best Participant} & \textbf{Single} & \textbf{Ensemble} & \textbf{Best Participant} & \textbf{Single} & \textbf{Ensemble} \\
\midrule
1 & 12.2\% (fa1) & 25.2\% & \textbf{29.2\%} & 16.8\% (fa1) & \textbf{29.1\%} & 27.9\% \\
2 & 22.6\% (LabZhu-FDU) & 16.4\% & \textbf{24.2\%} & 15.5\% (LabZhu-FDU) & \textbf{25.8\%} & 20.8\% \\
3 & 24.4\% (oaqa-3b-3) & \textbf{24.7\%} & 20.6\% & \textbf{48.3\%} (oaqa-3b-3) & 31.8\% & 33.3\% \\
4 & 32.5\% (oaqa-3b-4) & 34.0\% & \textbf{40.3\%} & \textbf{31.2\%} (oaqa-3b-4) & 29.0\% & 24.1\% \\
5 & \textbf{28.5\%} (oaqa-3b-5) & 23.7\% & 23.2\% & \textbf{29.0\%} (oaqa-3b-5) & 23.5\% & 26.1\% \\
\midrule
Avg. & 24.0\% & 24.8\% & \textbf{27.5\%} & \textbf{28.1\%} & 27.8\% & 26.5\% \\
\bottomrule
\end{tabularx}
  \caption{
  Comparison to systems on last year's (fourth) BioASQ challenge for factoid and list questions.
  For each batch and question type, we list the performance of the best competing system, our single model and ensemble.
  Note that our qualitative analysis (Section~\ref{sec:qualitative_analysis}) suggests that our factoid performance on batch 5 would be about twice as high if all synonyms were contained in the gold standard answers.
  }
  \label{tab:results_bioasq}
\end{table*}

Because the final results of the fifth BioASQ challenge are not available at the time of writing, we compare our system to the best systems in last year's challenge \footnote{Last year's results are available at \url{http://participants-area.bioasq.org/results/4b/phaseB/}}.
For comparison, we use the best single model and the model ensemble trained on \emph{BioASQ3B} (see Section~\ref{sec:ensemble}).
We then evaluate the model on the 5 batches of last year's challenge using the official BioASQ evaluation tool.
Each batch contains $100$ questions of which only some are factoid and list questions.
Note that the results underestimate our system's performance, because our competing system's responses have been manually evaluated by humans while our system's responses are evaluated automatically using string matching against a potentially incomplete list of synonyms.
In fact, our qualitative analysis in Section~\ref{sec:qualitative_analysis} shows that many answers are counted as incorrect, but are synonyms of the gold-standard answer.
The results are summarized in Table~\ref{tab:results_bioasq} and compared to the best systems in the challenge in each of the batches and question type categories.

With our system winning four out of five batches on factoid questions, we consider it state-of-the-art in biomedical factoid question answering, especially when considering that our results might be higher on manual evaluation.
The results on list questions are slightly worse, but still very competitive.
This is surprising, given that the network never saw a list question prior to the fine-tuning phase.
Due to small test set sizes, the sampling error in each batch is large, causing the single model to outperform the model ensemble on some batches.

\subsection{Qualitative Analysis}
\label{sec:qualitative_analysis}

In order to get a better insight into the quality of the predictions, we manually validated the predictions for the factoid questions of batch 5 of the fourth BioASQ challenge as given by the best single model (see Table~\ref{tab:results_bioasq}).
There are in total 33 factoid questions, of which 23 have as the gold standard answer a span in one of the contexts.
According to the official BioASQ evaluation, only 4 questions are predicted correctly (i.e., the gold standard answer is ranked highest).
However, we identified 10 rank-1 answers which are not counted as correct but are synonyms to the gold standard answer.
Examples include "CMT4D disease" instead of "Charcot-Marie-Tooth (CMT) 4D disease", "tafazzin" instead of "Tafazzin (TAZ) gene", and "$\beta$-glucocerebrosidase" instead of "Beta glucocerebrosidase".
In total, we labeled 14 questions as correct and 24 questions as having their correct answer in the top 5 predictions.

In the following, we give examples of mistakes made by the system.
Questions are presented in italics.
In the context, we underline predicted answers and present correct answers in boldface.

We identified eight questions for which the semantic type of the top answer differs from the question answer type.
Some of these cases are completely wrong predictions.
However, this category also includes subtle mistakes like the following:

\texttt{\textit{In which yeast chromosome does the rDNA cluster reside?}}

\texttt{The rDNA cluster in \underline{Saccharomyces cerevisiae} is located 450 kb from the left end and 610 kb from the right end of \textbf{chromosome XII}...}

Here, it predicted a yeast species the rDNA cluster is located in, but ignored that the question is asking for a chromosome.

Another type of mistakes is that the top answer is somewhat correct, but is missing essential information.
We labeled four predictions with this category, like the following example:

\texttt{\textit{How early during pregnancy does non-invasive cffDNA testing allow sex determination of the fetus?}}

\texttt{\textbf{Gold Standard Answer:} "6th to 10th week of gestation" or "first trimester of pregnancy"}

\texttt{\textbf{Given Top Answer:} "6th-10th"}

In summary, to our judgment, 14 of 33 questions ($42.4\%$) are answered correctly, and 24 of 33 questions ($72.7\%$) are answered correctly in one of the top 5 answers.
These are surprisingly high numbers considering low MRR score of $23.7\%$ of the automatic evaluation (Table~\ref{tab:results_bioasq}).

\section{Discussion and future work}

The most significant result of this work is that state-of-the-art results in biomedical question answering can be achieved even in the absence of domain-specific feature engineering.
Most competing systems require structured domain-specific resources, such as biomedical ontologies, parsers, and entity taggers. While these resources are available in the biomedical domain, they are not available in most domains.

Our system, on the other hand, requires a large open-domain QA dataset, biomedical word embeddings (which are trained in an unsupervised fashion), and a small biomedical QA dataset.
This suggests that our methodology is easily transferable to other domains as well.

Furthermore, we explored several supervised domain adaptation techniques.
In particular, we demonstrated the usefulness of forgetting cost for factoid questions.
The decreased performance on list questions is not surprising, because the model's performance on those questions is very poor prior to fine-tuning which is due to the lack of list questions in SQuAD.
We believe that large scale open-domain corpora for list questions would enhance performance further.

Unsupervised domain adaptation could be an interesting direction for future work, because the biomedical domain offers large amounts of textual data, some of which might even contain questions and their corresponding answers. We believe that leveraging these resources holds potential to further improve biomedical QA.

\section{Conclusion}

In this paper, we described a deep learning approach to address the task of biomedical question answering by using domain adaptation techniques.
Our experiments reveal that mere fine-tuning in combination with biomedical word embeddings yield state-of-the-art performance on biomedical QA, despite the small amount of in-domain training data and the lack of domain-dependent feature engineering.
Techniques to overcome catastrophic forgetting, such as a forgetting cost, can further boost performance for factoid questions.
Overall, we show that employing domain adaptation on neural QA systems trained on large-scale, open-domain datasets can yield good performance in domains where large datasets are not available.

\section*{Acknowledgments}

This research was supported by the German Federal Ministry of Education and Research (BMBF) through Software Campus project GeNIE (01IS12050).

\bibliography{paper}
\bibliographystyle{paper}

\end{document}